\documentclass[pdflatex,sn-aps,iicol]{sn-jnl} %% 开启双栏
\usepackage[table]{xcolor}   % 加载表格着色支持
\usepackage{graphicx}%
\usepackage{multirow}%
\usepackage{amsmath,amssymb,amsfonts}%
\usepackage{amsthm}%
\usepackage{mathrsfs}%
\usepackage[title]{appendix}%
\usepackage{xcolor}%
\usepackage{textcomp}%
\usepackage{manyfoot}%
\usepackage{booktabs}%
\usepackage{indentfirst}
\usepackage{cleveref}
\usepackage{algpseudocode}%
\usepackage{catchfilebetweentags}
\usepackage{listings}%
\usepackage[numbers]{natbib}%
\usepackage{array}
\usepackage{booktabs}
\usepackage{geometry}              % 空加载一次，不带选项
\usepackage{enumitem}         % 强大的列表定制宏包
\usepackage{siunitx}
\usepackage{threeparttable}
\usepackage{caption}   % 可选，用来微调 caption 样式
\usepackage{ragged2e} % 提供 \RaggedRight 等命令
\usepackage{array}    % 支持 >{...} 和 @{...} 的列格式定义
\usepackage{subfigure} % 支持子图
\theoremstyle{thmstyleone}%

\theoremstyle{thmstyletwo}%
\theoremstyle{thmstylethree}%
\usepackage[ruled,vlined]{algorithm2e}
\setcitestyle{sort&compress}
\begin{document}
\title[Article Title]{High-Frequency Semantics and Geometric Priors for End-to-End Detection Transformers in Challenging UAV Imagery}

\author[1,2]{\fnm{Hongxing} \sur{Peng}\textsuperscript{\textdagger}}\email{xyphx@scau.edu.cn}
\author[1]{\fnm{Lide} \sur{Chen}\textsuperscript{\textdagger}}\email{20243170009@stu.scau.edu.cn}
% --- 其他作者保持不变 ---
\author[3]{\fnm{Hui} \sur{Zhu}}\email{zhuhui@gdmec.edu.cn}
\author*[1]{\fnm{Yan} \sur{Chen}}\email{cheny@scau.edu.cn}
% --- Affiliations 保持不变 ---
\affil[1]{\orgdiv{College of Mathematics and Informatics}, \orgname{South China Agricultural University}, \city{Guangzhou}, \postcode{510642},  \country{China}}
\affil[2]{\orgdiv{Key Laboratory of Smart Agricultural Technology in Tropical South China, Ministry of Agriculture and Rural Affairs}, \city{Guangzhou}, \postcode{510642}, \country{China}}
\affil[3]{\orgdiv{School of Economics and Trade}, \orgname{Guangdong Mechanical \&Electrical Polytechnic}, \city{Guangzhou}, \postcode{510545}, \country{China}}

\abstract{Object detection in Unmanned Aerial Vehicle (UAV) imagery is fundamentally challenged by a prevalence of small, densely packed, and occluded objects within cluttered backgrounds. Conventional detectors struggle with this domain, as they rely on hand-crafted components like pre-defined anchors and heuristic-based Non-Maximum Suppression (NMS),  creating a well-known performance bottleneck in dense scenes. Even recent end-to-end frameworks have not been purpose-built to overcome these specific aerial challenges, resulting in a persistent performance gap. To bridge this gap, we introduce HEDS-DETR, a holistically enhanced real-time Detection Transformer tailored for aerial scenes. Our framework features three key innovations. First, we propose a novel High-Frequency Enhanced Semantics Network (HFESNet) backbone, which yields highly discriminative features by preserving critical high-frequency details alongside robust semantic context. Second, our Efficient Small Object Pyramid (ESOP) counteracts information loss by efficiently fusing high-resolution features, significantly boosting small object detection. Finally, we enhance decoder stability and localization precision with two synergistic components: Selective Query Recollection (SQR) and Geometry-Aware Positional Encoding (GAPE), which stabilize optimization and provide explicit spatial priors for dense object arrangements. On the VisDrone dataset, HEDS-DETR achieves a +3.8\% AP and +5.1\% AP$_{50}$ gain over its baseline while reducing parameters by 4M and maintaining real-time speeds. This demonstrates a highly competitive accuracy-efficiency balance, especially for detecting dense and small objects in aerial scenes.}

\keywords{Unmanned Aerial Vehicle , High-Frequency Feature Enhancement , Positional Prior Encoding , Real-time Detection , Dense Scene Understanding}
\maketitle

\begingroup
\renewcommand\thefootnote{\textdagger}
\footnotetext{These authors contributed equally: Hongxing Peng and Lide Chen.}
\endgroup

\section{Introduction}

Real-time object detection on Unmanned Aerial Vehicles (UAVs) has emerged as a cornerstone technology, underpinning a multitude of critical applications ranging from precision agriculture \cite{wang2024spatial} and highway surveillance \cite{zheng2024efficient} to urban environmental assessment \cite{henn2024surface}. However, the unique vantage point of aerial imagery presents a confluence of complex challenges that push the performance of existing detection models to their limits. Models must contend with high-density scenes populated with numerous small objects and frequent occlusions caused by cluttered backgrounds, all while operating under the stringent constraints of limited on-board computational resources and the demand for real-time processing \cite{tong2020recent}.

Prevailing UAV object detection models \cite{zhang2021vit,zhu2021tph,he2024key,peng2024lgff,yang2022querydet,xu2023dynamic,yang2019clustered, tang2024hic} rely on predefined anchor boxes and Non-Maximum Suppression (NMS), which introduce inherent limitations. Anchor box parameters require meticulous tuning for specific datasets and lack generalization capability. Meanwhile, NMS is sensitive to its threshold, potentially leading to the erroneous suppression of highly overlapping or adjacent objects or thEe failure to remove redundant detections. These issues create significant processing bottlenecks, particularly in the dense-object scenarios common in UAV footage.

The advent of the Detection Transformer (DETR) \cite{carion2020end} marked a paradigm shift towards truly end-to-end, NMS-free detection, though its substantial computational overhead initially rendered it impractical for real-time applications. A major breakthrough arrived with RT-DETR \cite{zhao2024detrs}, the first end-to-end framework to surpass state-of-the-art YOLO models under real-time conditions. Its advantage is credited to an innovative hybrid encoder design that decouples intra-scale feature interaction from cross-scale feature fusion.

\begin{figure}
\centering
\includegraphics[scale=0.36]{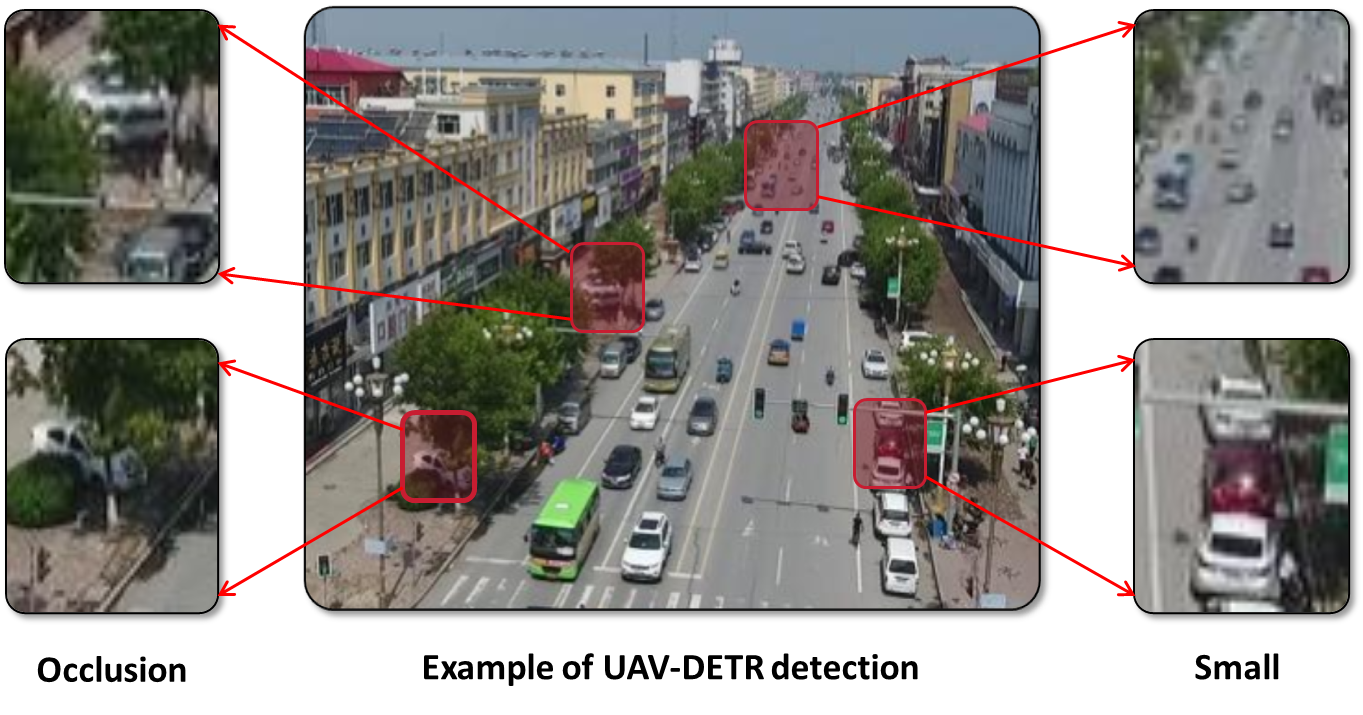}
\caption{Challenges in UAV Object Detection: dense small objects and cluttered backgrounds}\label{fig:visdrone}
\end{figure}

Despite RT-DETR's impressive performance on general-purpose benchmarks such as COCO \cite{lin2014microsoft}, its architecture is not inherently optimized for the distinct challenges of aerial imagery—namely, small target sizes, high object density, and frequent occlusions, as illustrated in \cref{fig:visdrone}. Its feature extractor and decoder were not purpose-built to address these core problems endemic to the aerial domain. Consequently, while RT-DETR provides a powerful baseline, unlocking its full potential for aerial object detection necessitates a targeted redesign of its core components. 

To address these gaps, we propose HEDS-DETR, a novel framework that systematically enhances RT-DETR for aerial scenes. First, to remedy inadequate feature representation, we replace the standard backbone with a High-Frequency Enhanced Semantics Network (HFESNet). This network is engineered to preserve fine-grained details crucial for distinguishing small, overlapping objects, while still capturing robust semantic context. Second, to combat information loss from aggressive downsampling, we introduce an Efficient Small Object Pyramid (ESOP). This novel feature pyramid efficiently integrates high-resolution S2 features, significantly boosting small object detection. Finally, we bolster the decoder's localization capabilities with a two-pronged strategy: a Selective Query Recollection (SQR) training mechanism mitigates cascading errors for more stable optimization, while a Geometry-aware Position Encoding (GAPE) provides explicit spatial priors to guide the precise localization of tightly clustered targets.

Our main contributions are summarized as follows:
\begin{enumerate}[label=\arabic*),, leftmargin=*, labelsep=1em, itemsep=1ex, align=left]
\item We propose the HFESNet, a novel backbone that yields more discriminative features for dense aerial scenes by optimally balancing the preservation of high-frequency spatial details and the extraction of high-level semantics.
\item We introduce the ESOP, a multi-scale fusion neck that strategically integrates high-resolution (S2) features via a lightweight pre-fusion module, significantly enhancing small-object detection with minimal computational overhead.
\item We stabilize the decoder and improve its localization accuracy via two synergistic innovations: SQR, a training strategy that mitigates error accumulation for stable bounding box refinement, and GAPE, a module that injects rich spatial priors to markedly improve precision for dense objects.
\item Our integrated model, HEDS-DETR, significantly outperforms its baseline on the VisDrone dataset \cite{zhu2021detection}. While maintaining real-time performance, it achieves a +5.1\% AP$_{50}$ and +3.8\% AP increase, with 17\% fewer parameters, thereby validating the efficacy of our targeted architectural redesign.
\end{enumerate}

\begin{figure*}
\centering
\includegraphics[width=1\textwidth]{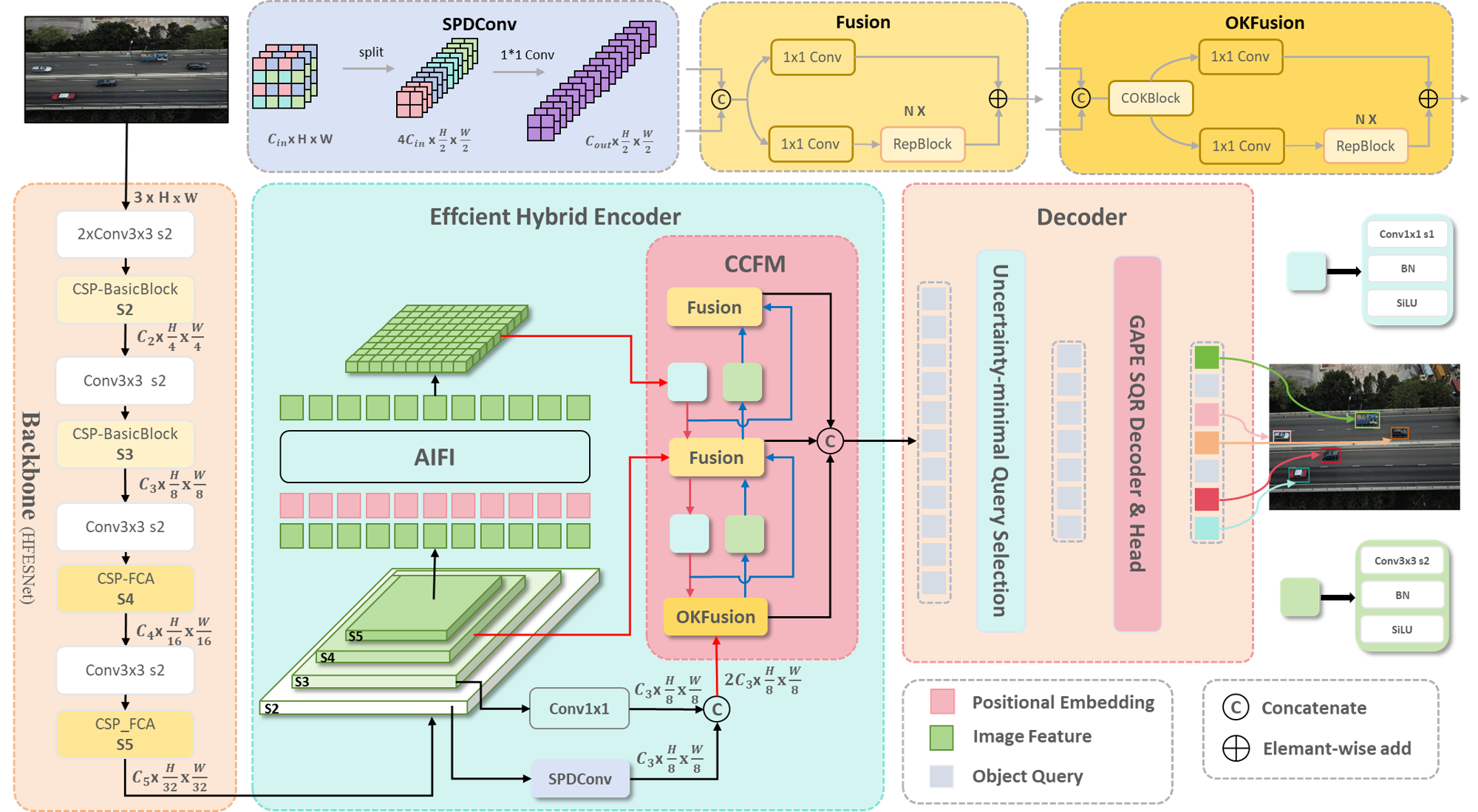}
\caption{Overview of HEDS-DETR.}\label{fig:overview}
\end{figure*}
\section{Related Work}

\subsection{Real-time and Aerial Object Detection}
Real-time object detection on Unmanned Aerial Vehicles (UAV-OD) is a challenging domain characterized by dense small objects, occlusions, and stringent computational constraints \cite{tong2020recent}. The YOLO series of models has long set a high standard for balancing speed and accuracy \cite{zhang2021vit,zhu2021tph,he2024key,peng2024lgff,yang2022querydet,xu2023dynamic,yang2019clustered, tang2024hic}. However, their reliance on hand-crafted Non-Maximum Suppression (NMS) introduces processing latency and hyperparameter sensitivity, which becomes a bottleneck in the dense-object scenarios typical of aerial views.

The advent of the Detection Transformer (DETR) \cite{carion2020end} introduced an end-to-end, NMS-free paradigm, but its high computational cost initially limited its real-time applicability. RT-DETR \cite{zhao2024detrs} marked a significant breakthrough, becoming the first real-time end-to-end detector to surpass leading YOLO models. Its efficiency stems from a hybrid encoder that decouples intra-scale interaction from cross-scale fusion. Despite its strong performance on general datasets like COCO \cite{lin2014microsoft}, RT-DETR's architecture is not inherently optimized for the distinct challenges of aerial imagery, such as pervasive small targets and high object density. This necessitates targeted adaptations to its core components to unlock its full potential for UAV-OD.

\subsection{Architectural Enhancements for Aerial Detection}
Optimizing detectors for aerial scenes requires targeted modifications to both feature representation and decoder design to enhance the handling of small, dense, and occluded objects.

Feature Representation and Fusion: Effective discrimination in aerial scenes requires features that are both semantically rich and spatially precise. While some modules aim to enhance semantics, such as FGE \cite{huang2024discriminative} for small objects or CSP-COT \cite{gao2024oriented} for global context, they risk suppressing the high-frequency details (e.g., edges, textures) vital for distinguishing overlapping objects. Concurrently, standard multi-scale fusion strategies, from FPN \cite{lin2017feature} to its successors like PANet \cite{liu2018path} and BiFPN \cite{tan2020efficientdet}, often discard the highest-resolution feature maps for computational efficiency. RT-DETR follows this trend by omitting the S2 feature stage, leading to an irreversible loss of fine-grained spatial information critical for detecting minute objects. Our work addresses these dual challenges by developing a backbone (HFESNet) that explicitly fuses high-frequency details with deep semantic context, and an efficient feature pyramid (ESOP) that reintegrates high-resolution features without compromising real-time performance.

Decoder Stability and Precision: The performance of transformer-based decoders can be hindered by two key issues: cascading prediction errors and insufficient spatial priors for localization. Error accumulation, where inaccuracies from early decoding stages propagate and degrade subsequent refinement, is a known problem \cite{sun2021sparse}, especially in cluttered aerial scenes. Research like SQR-DETR \cite{chen2023enhanced} has shown that refining queries with stronger supervision from earlier stages can effectively mitigate this. Furthermore, precise localization of dense objects demands rich spatial guidance. The standard RT-DETR decoder's reliance on a simple linear projection of reference points provides weak spatial constraints. While prior works such as PETR \cite{liu2022petr} and DAB-DETR \cite{liu2022dab} have demonstrated the benefits of more explicit geometric or conditional spatial priors, an efficient formulation for real-time aerial detection remains an open challenge. To this end, we introduce a selective query recollection (SQR) training strategy to stabilize optimization and a geometry-aware position encoding (GAPE) to provide the decoder with explicit spatial priors, enhancing localization accuracy for dense targets.

\begin{figure*}
\centering
\includegraphics[width=\textwidth]{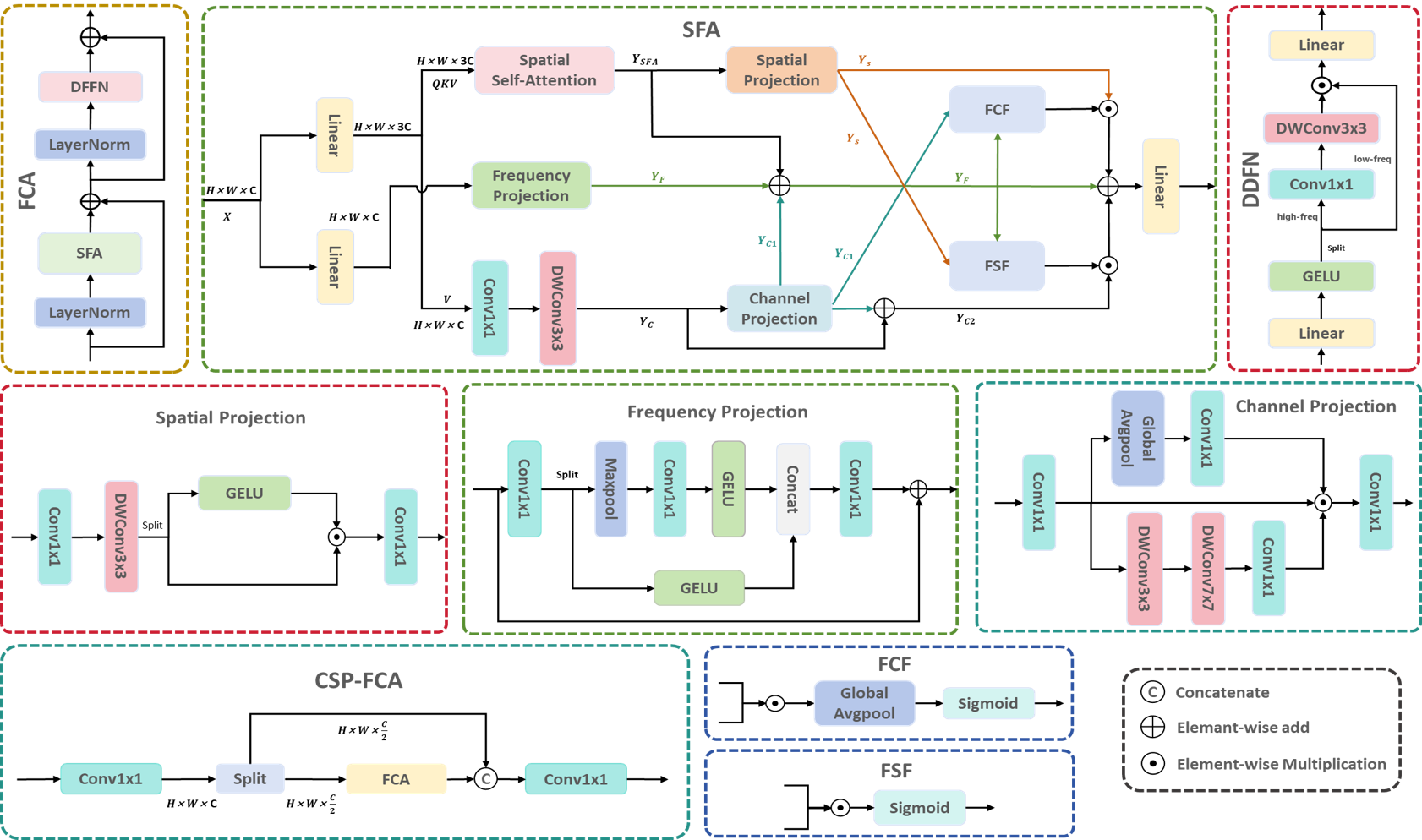}
\caption{The architecture of our proposed CSP-FCA module, which integrates the efficiency of the CSP strategy with the detail-recovery capabilities of the FCA block.}\label{fig:csp-fca}
\end{figure*}

\section{Methodologies}\label{sec3}
As illustrated in \cref{fig:overview}, this study proposes HEGS-DETR based on the RT-DETR-R18 architecture \cite{zhao2024detrs}. We enhance the model through four critical components/strategies. First, we develop HEFSNet to replace the original backbone network, which facilitates discriminative feature learning for improved object separability under scenarios with dense small targets and complex backgrounds. Second, we propose the ESOP strategy to balance model complexity while effectively enhancing small object detection performance. Third, we introduce the GAPE strategy to construct explicit spatial priors that strengthen the model's capability for precise localization. Finally, we design the SQR training strategy to stabilize bounding box optimization by shifting the training focus backward and alleviating decoder-induced error accumulation.

\subsection{High-Frequency Enhanced Semantics Network}

While deep semantic features are vital for discriminating objects in complex scenes, conventional backbones inherently suppress the high-frequency details (e.g., edges, textures) necessary for resolving dense and occluded targets. To address this trade-off, we propose the High-Frequency Enhanced Semantics Network (HFESNet) as a new backbone architecture. HFESNet is engineered to enrich deep semantic features with fine-grained spatial information by strategically replacing standard residual blocks in ResNet-18 with our novel CSP-FCA and CSP-BasicBlock modules.

The central component of HFESNet is our proposed CSP-FCA module, illustrated in \cref{fig:csp-fca}. This module synergizes the efficiency of the Cross Stage Partial (CSP) design \cite{wang2020cspnet} with the detail-recovery capabilities of Frequency-aware Cascade Attention (FCA) \cite{freqformer2024}. Following the CSP principle, the input features $\mathbf{X}$ are partitioned into two parallel branches (\cref{eq:1,eq:2}). One branch preserves a portion of the original features, which optimizes gradient flow and reduces computational load. The other branch is processed by the FCA module, which is specifically designed to recover and enhance high-frequency information often lost in deep layers. The two branches are then concatenated and fused, as formulated in \cref{eq:3}, yielding an output that is both semantically rich and spatially precise.

\begin{align}
\mathbf{X}' &= \text{Conv}_{1 \times 1}(\mathbf{X}) \in \mathbb{R}^{H \times W \times C}, \label{eq:1} \\
\mathbf{X}' &\xrightarrow[\text{split}]{\text{channel}}
\begin{cases}
\mathbf{X}_1' \in \mathbb{R}^{H \times W \times \frac{C}{2}}, \\
\mathbf{X}_2' \in \mathbb{R}^{H \times W \times \frac{C}{2}},
\end{cases} \label{eq:2} \\
\mathbf{Y} &= \text{Conv}_{1 \times 1}(\text{Concat} (\mathbf{X}_1', \mathcal{F}_{\text{FCA}}(\mathbf{X}_2'))). \label{eq:3}
\end{align}

The efficacy of our design stems from the FCA module, which replaces standard self-attention. As shown in \cref{fig:csp-fca}, FCA employs two key mechanisms: 1) Spatial Frequency Attention (SFA), which jointly models features across spatial, channel, and frequency domains to capture comprehensive dependencies, and 2) a Dual Frequency Fusion Feed-Forward Network (DFFN), which explicitly separates and enhances high-frequency components before fusing them with low-frequency semantic information. The overall transformation, governed by residual connections (\cref{eq:4,eq:5}), ensures that our backbone retains robust semantic context while recovering critical high-frequency details lost during downsampling.

\begin{align}
X_{SFA} &= \text{SFA}(\text{LayerNorm}(X_{in})) \oplus X_{in} \label{eq:4} \\
X_{out} &= \text{DDFN}(\text{LayerNorm}(X_{SFA})) \oplus X_{SFA} \label{eq:5}
\end{align}

where $\oplus$ denotes element-wise addition.

\subsection{Efficient Small Object Pyramid}
\begin{figure*}
\centering
\includegraphics[width=1\textwidth]{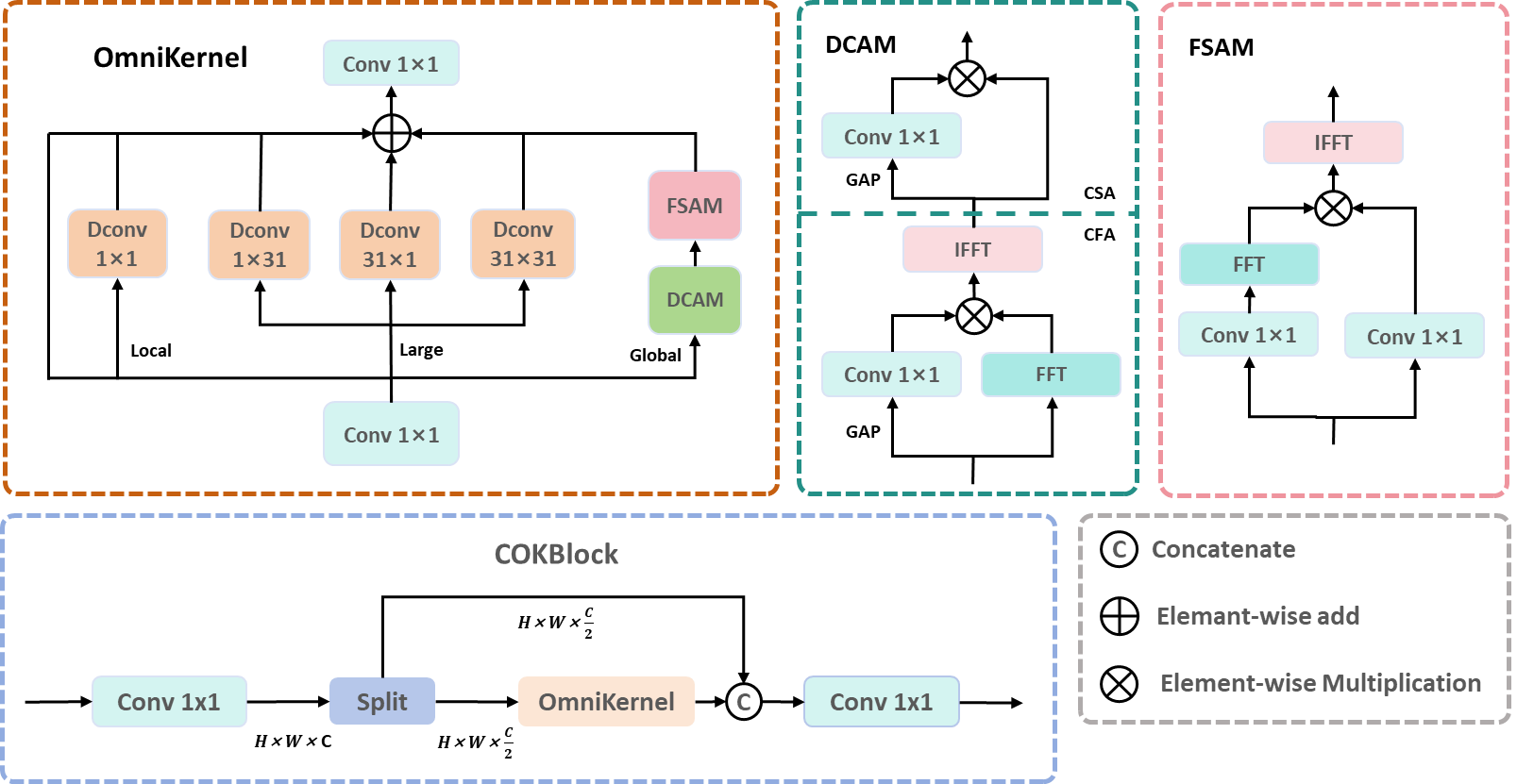}
\caption{The architecture of the proposed Cross-scale Omni-Kernel Block (COKBlock). FFT, IFFT, and GAP denote the Fourier Transform, Inverse Fourier Transform, and Global Average Pooling, respectively.}\label{fig:cokblock}
\end{figure*}

The reliance of RT-DETR on low-resolution feature maps from its final three stages impairs its ability to detect small objects. While incorporating higher-resolution features from early layers can provide crucial detail, it introduces significant computational overhead. To address this, we propose the Efficient Small Object Pyramid (ESOP) strategy, which efficiently integrates multi-scale features without prohibitive costs.

Our approach begins by generating a detail-rich feature map, $S_{23}$, by processing the early-stage features $S_2$ and $S_3$. We use SPDConv \cite{sunkara2022no} in place of strided convolution on $S_2$ to preserve fine-grained spatial information critical for small object detection. This is then concatenated with the processed $S_3$ feature map.
\begin{equation}
S_{23} = \text{Concat}(\text{SPDConv}(S_2), \text{Conv}(S_3))
\end{equation}
However, naively fusing $S_{23}$ with the deep, semantic-rich feature map $S_{45}$ (derived from stages $S_4$ and $S_5$) is challenging due to their disparate scales and semantic levels, which can lead to feature conflicts. To bridge this semantic gap, we introduce the OKFusion module, which is centered around a novel Cross-scale Omni-Kernel Block (COKBlock). This process is formulated as:
\begin{align}
S' &= \text{COKBlock}(\text{Concat}(S_{23}, S_{45})) \\
S_{2345} &= \text{Conv}(S') \oplus \text{RepBlock}(\text{Conv}(S'))
\end{align}
As illustrated in \cref{fig:cokblock}, the COKBlock leverages a CSP architecture and an OmniKernel module \cite{cui2024omni} to perform comprehensive feature fusion. The OmniKernel parallelly extracts multi-scale context from the input features $X_{in}$ via three specialized branches:
\begin{itemize}
    \item \textbf{Large-Kernel Branch ($X_{Large}$):} Employs large (31$\times$31) and strip (1$\times$31, 31$\times$1) depth-wise convolutions to explicitly model long-range spatial dependencies.
    \item \textbf{Global-Context Branch ($X_{Global}$):} Operates in the frequency domain, using a Dual-domain Channel Attention Module (DCAM) and a Frequency Space Attention Module (FSAM) \cite{cui2024omni} to capture global context by enhancing information-rich frequency components.
    \item \textbf{Local-Feature Branch ($X_{Local}$):} Utilizes a 1$\times$1 depth-wise convolution to modulate and preserve fine-grained local signals.
\end{itemize}
The outputs of these branches are aggregated to form a unified, multi-scale representation:
\begin{align}
X_{Large} &= 
  \mathrm{DConv}_{1\times31}(X_{in})
  + \mathrm{DConv}_{31\times1}(X_{in}) \notag\\
&\quad +\, \mathrm{DConv}_{31\times31}(X_{in}) \label{eq:xl}\\[2pt]
X_{Global} &= \mathrm{FSAM}\!\left(\mathrm{DCAM}(X_{in})\right) \label{eq:xg}\\[2pt]
X_{Local}  &= \mathrm{DConv}_{1\times1}(X_{in}) \label{eq:xlc}\\[2pt]
X_{out} &= X_{in} + X_{Local} + X_{Large} + X_{Global} \label{eq:xout}
\end{align}

where DConv denotes depth-wise convolution. This design enables the COKBlock to effectively fuse features from global to local scales, enhancing the model's small object detection capabilities with high computational efficiency.

\subsection{Geometry-Aware Position Encoding}
The standard RT-DETR decoder employs a linear projection of reference points for positional encoding, which is suboptimal for capturing the intricate geometric relationships between object queries and multi-scale image features. This limitation can impair localization precision, especially for small or dense objects. To address this, we introduce Geometry-Aware Position Encoding (GAPE), a method that constructs explicit spatial priors by synergizing reference box geometry with content query information.

\begin{figure}
\centering
\includegraphics[scale=0.53]{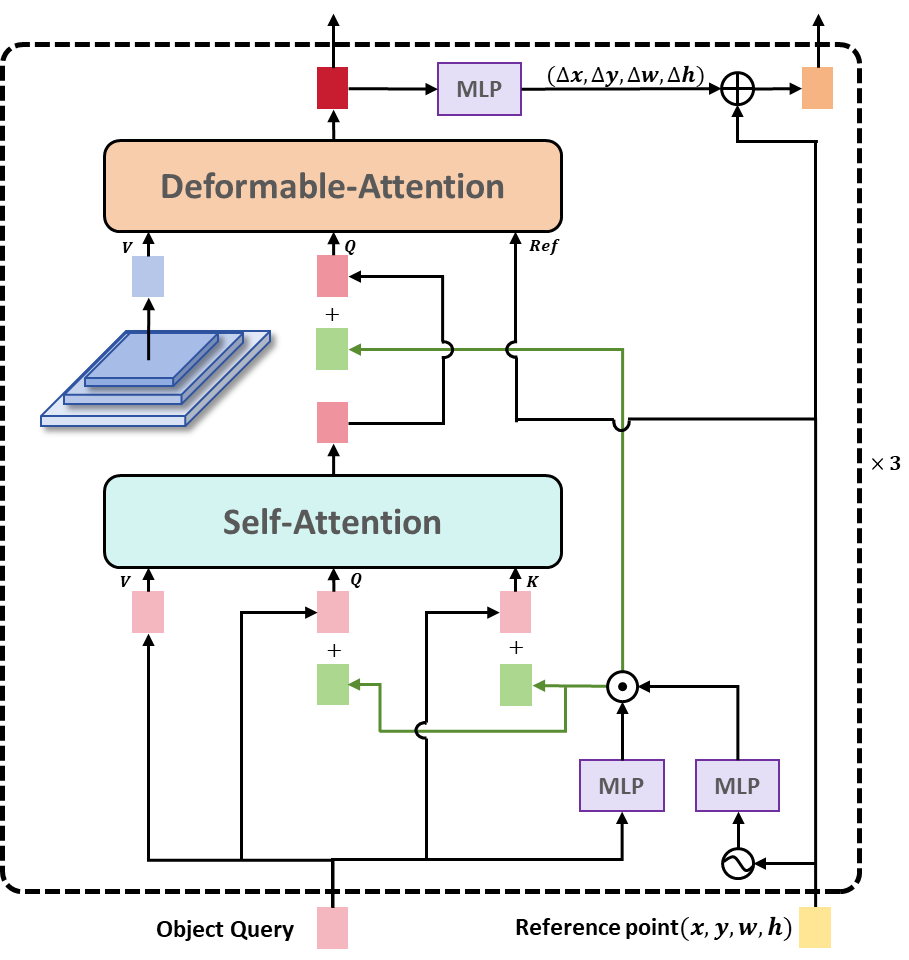}
\caption{Illustrates the GAPE strategy at decoder. The small green rectangles in the diagram denote the positional encodings we designed.}\label{fig:GAPE}
\end{figure}

As illustrated in \cref{fig:GAPE}, GAPE operates in two stages. First, it generates a positional query $P_q$ for each object query. The four coordinates of the reference box $B_q=(x_q, y_q, w_q, h_q)$ are individually encoded using sinusoidal functions and then concatenated. This high-dimensional geometric representation is processed by an MLP to form a base positional embedding. Concurrently, the content query $O_q$ is passed through another MLP to generate a dynamic scaling vector. The final positional query $P_q$ is obtained by element-wise multiplication of the base embedding and the scaling vector, as formulated below:
\begin{align}
    PE(B_{q}) &= \text{Cat}(\text{PE}(x_{q}), \text{PE}(y_{q}), \text{PE}(w_{q}), \text{PE}(h_{q})) \tag{13} \\
    P_{q} &=  \text{MLP}(O_{q}) \odot \text{MLP}(\text{PE}(B_{q})) \tag{14}
\end{align}
where $\odot$ denotes element-wise multiplication.

Second, we inject these geometry-aware priors $P_q$ into the decoder's attention mechanisms. For self-attention, $P_q$ is added to both queries and keys to model inter-query spatial relationships:
\begin{align}
\text{Self-Attn}: \quad
Q_{q} &= O_{q} + P_{q}, \quad
K_{q} = O_{q} + P_{q}, \notag\\ &\quad V_{q} = O_{q}
\tag{15}
\end{align}

For cross-attention, the GAPE-enhanced query $O_q' + P_q$ (where $O_q'$ is the output from self-attention) is used to generate the sampling offsets for the deformable attention mechanism:
\begin{align}
    \text{Cross-Attn} : Q_{q} = O_{q}' + P_{q}, \quad V = \text{Sampling}(S)
\tag{16}
\end{align}

This explicitly guides the sampling process to focus on salient object regions within the encoder's feature maps $S$, thereby significantly enhancing the model's localization capability for challenging scenarios like small and occluded objects.

\subsection{Selective Query Recollection}\label{sec:subsec3-4}

\begin{figure*}
\centering
\includegraphics[scale=0.5]{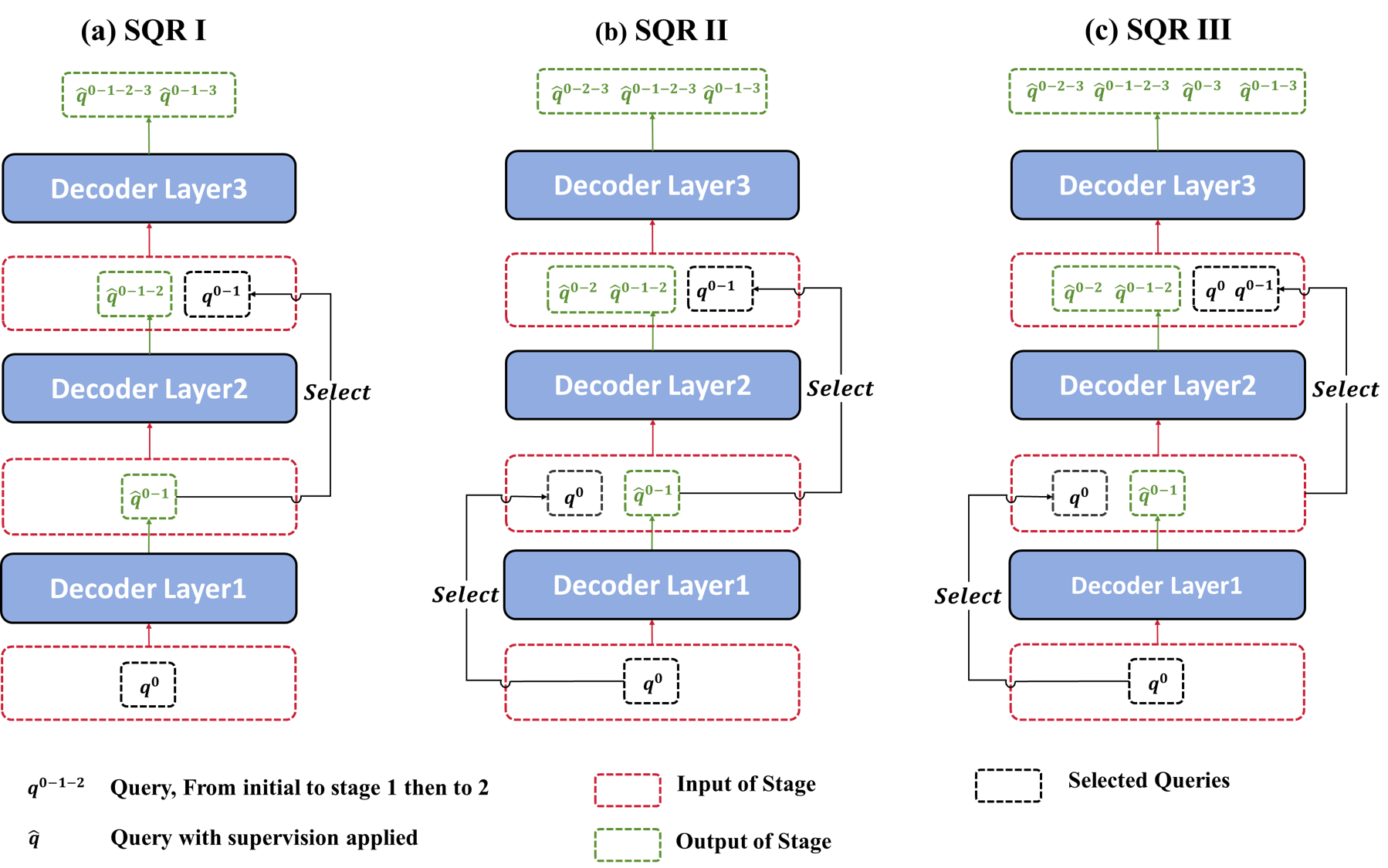}
\caption{Illustration of the three variants of our SQR.}\label{fig:sqr}
\end{figure*}

\begin{figure}
\centering
\includegraphics[scale=0.35 ]{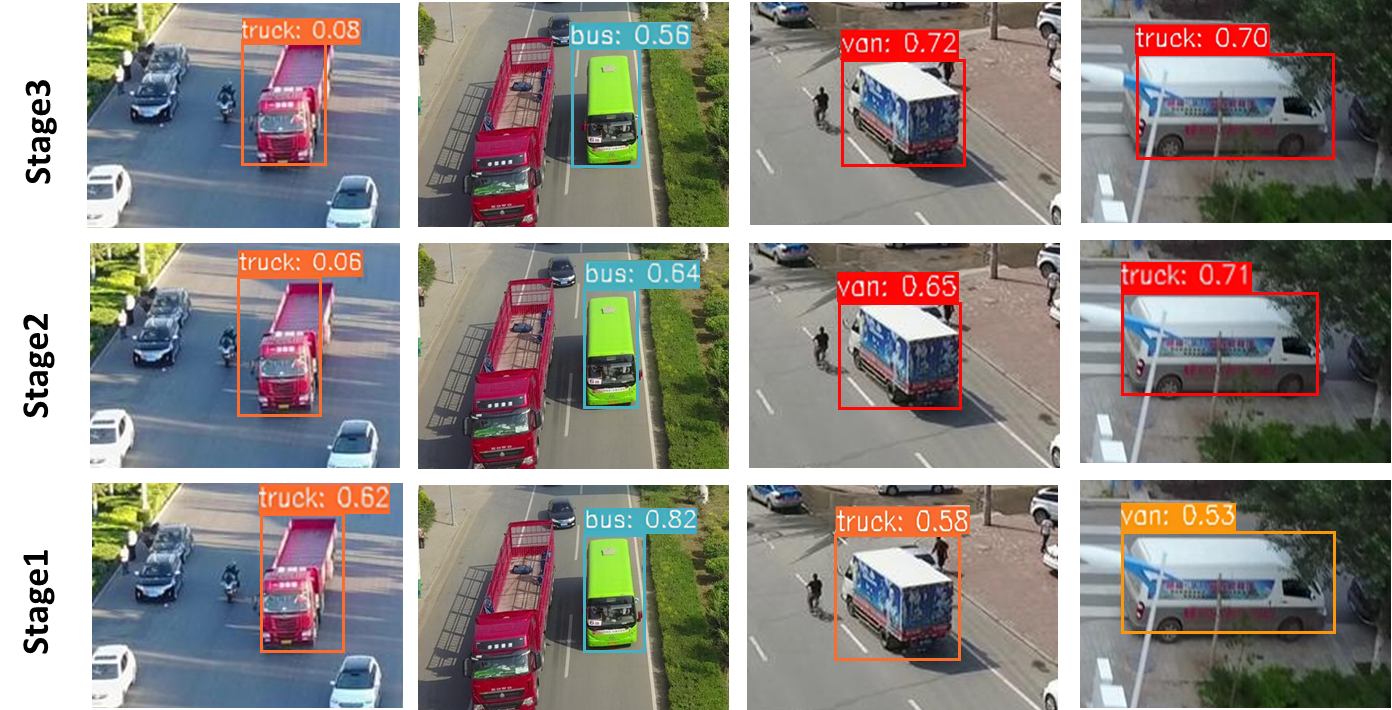}
\caption{Visualizes RT-DETR-R18 predictions across stages (bottom→top: early→late). The first two columns highlight confidence drops on initially correct detections; the last two show early correct classifications turning into errors later (red boxes).}\label{fig:tpfp}
\end{figure}

The progressive refinement strategy, implemented via multi-stage decoders in transformer-based object detectors, can paradoxically lead to performance degradation in later stages. This issue is particularly acute in challenging scenarios like UAV-based aerial imagery, which is often characterized by dense small objects and cluttered backgrounds. As shown in \cref{fig:tpfp}, this degradation manifests in two primary ways: (1) confidence fading, where the confidence of correct early-stage detections is erroneously reduced, leading to missed targets (false negatives); and (2) error amplification, where initial misclassifications become entrenched and solidified in subsequent stages (false positives). For instance, \cref{fig:tpfp} illustrates an early, correct detection being lost due to a severe confidence drop, and another correct detection being persistently misclassified in later stages.

We attribute this performance decay to two factors. First, the inherent cascaded architecture propagates both beneficial and detrimental query refinements downstream. Second, the final decoder stages, which have a decisive impact on the output, lack a mechanism to revisit less corrupted query states from earlier stages to correct inherited errors.

\begin{table}
\centering
\begin{tabular}{l|cc|c}
\toprule
\textbf{Stage} & \textbf{1} & \textbf{2} & \textbf{1\&2} \\ 
\midrule
TP F Rate (\%) & 19.87 & 18.95 & 33.84 \\ 
FP E Rate (\%) & 53.67 & 54.58 & 71.84 \\ 
\bottomrule
\end{tabular}
\caption{TP/FP Rate Analysis of RT-DETR-18’s Stage 3 with Single and Combined Layers at IoU \textgreater \  0.5}
\label{tab:sqr-tpfp-r18}
\end{table} 

To quantify this phenomenon, we adopt the True Positive Fading Rate (TP F Rate) and False Positive Exacerbation Rate (FP E Rate) metrics \cite{chen2023enhanced}. TP F Rate measures the percentage of true positives from an early stage that become false negatives in the final stage, while FP E Rate tracks error amplification. As detailed in \autoref{tab:sqr-tpfp-r18}, when comparing the final Stage 3 predictions against Stage 1, the TP F Rate is a significant 19.87\%. When evaluated against the union of correct predictions from Stages 1 and 2, the TP F Rate escalates to 33.84\% and the FP E Rate reaches 71.84\%. These statistics underscore the severity of error propagation in multi-stage decoding for complex scenes.

To address this issue,as shown in \cref{fig:sqr} we design three Selective Query Recollection (SQR) methods. Let $D^j$ denote the $j$-th decoder mapping. For any finite index sequence $\{k_1, k_2, \ldots, k_m\}$, we define:

\begin{equation}
q^{(0-k_1-k_2-\cdots-k_m)} := D^{k_m}\left(D^{k_{m-1}}\left(\cdots D^{k_1}(q^0)\cdots\right)\right) \tag{17}
\end{equation}

where $q^0$ represents the initial state. The query $q$ encompasses both object queries and reference points.

\textbf{SQR I}: This variant injects the input of the second decoder layer ($C^2$) into the input of the third layer ($C^3$). The aim is to provide the final, decisive layer with a less corrupted query set from an intermediate stage, reducing accumulated errors with minimal overhead.
\begin{align}
C^1 &= \{q^0\} \nonumber \\
C^2 &= D^1(C^1) = \{q^{0-1}\} \nonumber \\
C^3 &= D^2(C^2) \cup C^2 = \{q^{0-1-2}, q^{0-1}\} \nonumber \tag{18}
\end{align}

where $C^j$ represents the input to the $j$-th decoder layer, and the output of $D^j$ serves as the supervision signal for the $j$-th layer.

\textbf{SQR II}: SQR II augments the second ($D^2$) and third ($D^3$) decoders with initial ($q^0$) and first-pass refined ($q^{0-1}$) queries, respectively. This strategy aims to mitigate cascading errors by leveraging the potential correctness of initial queries to revise intermediate predictions.

\begin{align}
C^1 &= \{q^0\} \nonumber \\
C^2 &= D^1(C^1) \cup C^1 = \{q^{0-1}, q^0\} \nonumber \\
C^3 &= D^2(C^2) \cup D^1(C^1) = \{q^{0-1-2}, q^{0-2}, q^{0-1}\} \tag{19}
\end{align}

\textbf{SQR III}: As the most comprehensive variant, SQR III forwards the entire enriched input of $D^2$ (i.e., $\{q^{0-1}, q^0\}$) to become part of the input for $D^3$. This strategy maximally increases the diversity of antestage queries for the final layer, empowering it to perform more robust optimization by considering states from multiple previous points.

\begin{align}
C^1 &= \{q^0\}  \nonumber \\
C^2 &= D^1(C^1) \cup C^1 = \{q^{0-1}, q^0\}  \nonumber \\
C^3 &= D^2(C^2) \cup C^2 = \{q^{0-1-2}, q^{0-2}, q^{0-1}, q^0\} \tag{20}
\end{align}

Our empirical results show that SQR II consistently outperforms the other variants across all metrics. Consequently, we adopt SQR II as our final method, hereafter referred to as SQR. This approach effectively reduces both the TP F Rate and FP-ER, thereby stabilizing the progressive refinement of bounding boxes throughout the decoding process.

\section{Experiment}\label{sec4}
\subsection{Dataset and Experiment Setup}

\indent \textbf{Dataset:} In the domain of object detection in drone-captured aerial imagery, a crucial and widely adopted dataset is the VisDrone dataset \cite{zhu2021detection}. This dataset is partitioned into three subsets with a ratio of 7:2:1, comprising a training set (6471 images), a validation set (548 images), and a test set (1610 images). The images within this dataset possess a maximum resolution of $2000 \times 1500$ pixels and encompass 10 distinct object categories, including pedestrians, various vehicle types, bicycles, and motorcycles, thereby providing a comprehensive benchmark platform for the training and evaluation of object detection algorithms.

In addition, to evaluate the generalization ability of our model, we conducted experiments on the CARPK \cite{hsieh2017dronebasedobjectcountingspatially} dataset. The CARPK dataset consists of 1,448 images extracted from high-resolution videos recorded in four different parking lots, including 989 training images and 459 validation images. Unlike VisDrone, CARPK is characterized by a single class of dense small objects (vehicles), which makes it highly suitable for assessing the robustness of detection models in crowded scenarios.

\textbf{Implementation Details}: The experiments were conducted on a high-performance computing server equipped with dual NVIDIA GeForce RTX 8000 GPUs, operating on the Ubuntu 20.04 operating system. The deep learning framework utilized was PyTorch 2.5.0, accelerated with CUDA 12.2. We employed the Adam optimizer with an initial learning rate of $1 \times 10^{-4}$ and a weight decay of $0.0001$ to mitigate overfitting. The model was trained for a total of 300 epochs with a batch size of 16, processing input images at a resolution of $640 \times 640$. Momentum was set to 0.9 to expedite convergence during the optimization process. For the CARPK dataset, we maintained the same experimental setup but trained the model for 200 epochs. For comprehensive evaluation, we adopted the evaluation metrics from the COCO dataset, including Average Precision (AP) with Intersection over Union (IoU) thresholds ranging from 0.5 to 0.95 in increments of 0.05, as well as AP at different scales\cite{lin2014microsoft}.

\begin{figure*}
\centering
\includegraphics[scale=0.53]{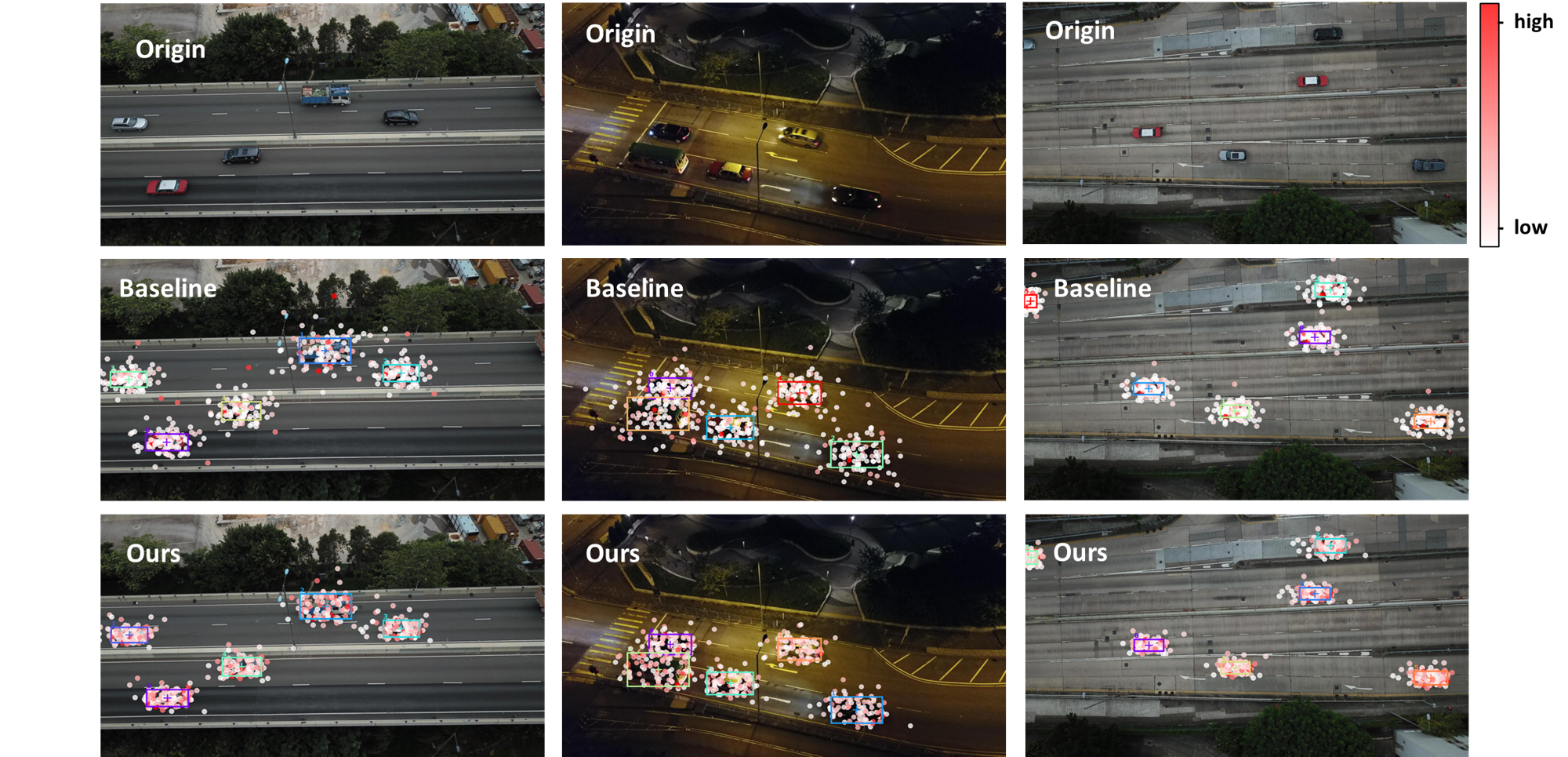}
\caption{Visualization of deformable attention in the decoder. Reference points are marked with a cross, and each sampling point is shown as a solid circle; redder colors indicate higher attention weights.}\label{fig:deform-att}
\end{figure*}

\subsection{Comparative Experiment}
We conducted extensive comparative experiments on the VisDrone-2019 validation set to evaluate our model's performance. As presented in Table \ref{tab:contrast}, HEDS-DETR is benchmarked against a diverse range of detectors, including traditional, YOLO-series, UAV-specific, and other end-to-end models.
The results reveal clear architectural trade-offs among existing methods. Traditional detectors (e.g., Faster R-CNN) and the popular YOLO series show limited accuracy on this challenging dataset, indicating their architectures struggle with dense, small objects in aerial scenes. Specialized UAV detectors offer mixed results; for instance, ClusDet achieves a high AP$_{50}$ but at a prohibitive computational cost (207 GFLOPS), highlighting a classic accuracy-efficiency dilemma. While recent end-to-end models like DEIM show promise, they do not consistently excel across all metrics.
In contrast, our HEDS-DETR demonstrates highly competitive performance. It significantly outperforms its baseline, RT-DETR-R18, by +3.8\% AP and +5.1\% AP$_{50}$ with only a marginal increase in GFLOPS. More importantly, HEDS-DETR surpasses all listed competitors on the most critical metrics for this task, achieving 29.4\% AP, 20.9\% AP$_s$, and 40.2\% AP$_m$. This demonstrates its exceptional capability in handling the most challenging small and medium-sized objects. In addition to its strong performance, our model is also more parameter-efficient, requiring 14\% fewer parameters than DEIM (16.53M vs. 19.19M). These comprehensive results validate that HEDS-DETR sets a new and superior trade-off between performance and efficiency for object detection in aerial imagery.

\begin{table*}
  \centering
  \caption{Comparative Experiments on the VisDrone-2019 Benchmark (Validation Set).}
  \label{tab:contrast}
  % 将 \textwidth 修改为 0.9\textwidth 或其他你希望的比例
  \resizebox{0.8\textwidth}{!}{%
    \begin{tabular}{l c c c c c c}
      \toprule
      \textbf{Model} 
        & \textbf{AP\textsubscript{50} (\%)} 
        & \textbf{AP\textsubscript{50--90} (\%)} 
        & \textbf{AP\textsubscript{s} (\%)} 
        & \textbf{AP\textsubscript{m} (\%)} 
        & \textbf{GFLOPS} 
        & \textbf{Params (M)} \\
      \midrule
      \rowcolor{gray!20}
      \multicolumn{7}{l}{\textbf{Traditional Detectors}} \\ 
      Faster RCNN \cite{ren15fasterrcnn}     & 39.6          & 24.3  & 15.1 & 36.6          & 208    & 41.39 \\
      Cascade-RCNN \cite{cai2018cascade}    & 38.7          & 24.2  & 15.1 & 36.5          & 236    & 69.29 \\
      ATSS \cite{zhang2020bridging}            & 39.6          & 25.2  & 14.9 & 38.9          & 110    & 38.91 \\
      TOOD \cite{feng2021tood}            & 41.1          & 25.6  & 15.9 & 38.1          & 199    & 32.04 \\
      \midrule
      \rowcolor{gray!20}
      \multicolumn{7}{l}{\textbf{YOLO Series Detectors \cite{redmon2016lookonceunifiedrealtime} }} \\ 
      YOLOv5m         & 39.4          & 23.6  & 14.0 & 35.7          & 64.0   & 25.05 \\
      YOLOv6m         & 38.8          & 23.4  & 13.0 & 36.1          & 161  & 51.98 \\
      YOLOv8m          & 40.3          & 24.3  & 14.9 & 36.7          & 79.3   & 25.09 \\
      YOLOv9m         & 42.0          & 25.1  & 15.6 & 37.4          & 77.9   & 20.22 \\
      YOLOv10m      & 40.0          & 24.0  & 14.2 & 35.9          & 58.9   & 15.32 \\
      YOLOv11m         & 41.5          & 25.2  & 15.9 & 37.0          & 68.5   & 20.11 \\
      YOLOv12m        & 41.2          & 24.8  & 15.4 & 37.0          & 60.4   & 19.67 \\
      \midrule
      \rowcolor{gray!20}
      \multicolumn{7}{l}{\textbf{UAV-based Object Detectors}} \\
      HIC-YOLOv5 \cite{tang2024hic}     & 44.3          & 26.0  & -    & -             & \textbf{31.2}  & \textbf{9.4} \\
      QueryDet  \cite{yang2022querydet}      & 48.1         & 28.3 & -    & -             & 212  & 33.9  \\
      DCFL  \cite{xu2023dynamic}         & 36.1          & -     & -    & -             & 157.8   & 32.1 \\
      ClusDet \cite{yang2019clustered}        & \textbf{50.6} & 26.7  & 17.6 & 38.9          & 207  & 30.2  \\
      \midrule
      \rowcolor{gray!20}
      \multicolumn{7}{l}{\textbf{End-to-End Detectors}} \\ 
      Deformable-DETR \cite{zhu2020deformable} & 43.7          & 25.6  & 17.3 & 36.1          & 193  & 40.78 \\
      DAB-Deformable-DETR \cite{liu2022dab}        & 44.2          & 25.8  & 18.1 & 35.5          & 260 & 47.06 \\
      RT-DETR-R18 \cite{zhao2024detrs}     & 42.9          & 25.6  & 17.6 & 35.4          & 57.0   & 19.88 \\  
      DFINE \cite{peng2024dfineredefineregressiontask}           & 43.6          & 26.1  & 18.3 & 35.6          & 56.4   & 19.19 \\        
      DEIM \cite{huang2024deim}            & 45.6          & 28.5  & 19.7 & 36.8          & 56.4   & 19.19 \\  
      HEDS-DETR (ours)& 48.0          & \textbf{29.4} & \textbf{20.9} & \textbf{40.2} & 64.9   & 16.53 \\
      \bottomrule
    \end{tabular}%
   }
\end{table*}

\subsection{Ablation Studies}

\begin{table*}[ht]
  \centering
  \caption{Ablation Study on Architectural Components of HEDS-DETR.}
  \label{tab:ablation}
  \begin{tabular}{cccc|ccc|ccc}
    \toprule
    HFESNet & ESOP & GAPE & SQR
      & AP$_{50}$ (\%) & AP$_{50\text{-}90}$ (\%) & AP$_s$ (\%)
      & GFLOPS & Params (M) & FPS \\
    \midrule
    – & – & – & –
      & 42.9 & 25.6 & 17.6
      & 57.0    & 19.88      & \textbf{172.4}  \\
    \checkmark & – & – & –
      & 45.7 & 27.6 & 19.4
      & \textbf{52.6}   &  \textbf{15.42}    & 158.7     \\
    \checkmark & \checkmark & – & –
      & 46.2 & 28.0 & 19.9
      & 64.6    & 16.30      & 135.1     \\
    \checkmark & \checkmark & \checkmark & –
      & 47.3 & 28.6 & 20.7
      & 64.9    & 16.53      & 131.6     \\
    \checkmark & \checkmark & \checkmark & \checkmark
      & \textbf{48.0} & \textbf{29.4} & \textbf{20.9}
      & 64.9    & 16.53      & 131.6     \\
    \bottomrule
  \end{tabular}
\end{table*}

To validate the efficacy of our proposed components, we conduct a comprehensive ablation study on the VisDrone validation set. All inference speeds (FPS) are benchmarked on a single NVIDIA RTX 3090 GPU. The results are summarized in Table \ref{tab:ablation}.
We start with the RT-DETR-R18 baseline, which obtains 42.9\% AP$_{50}$ and 17.6\% AP$_s$, establishing the challenge of small object detection in UAV imagery. First, by integrating HFESNet as the new backbone, AP$_{50}$ improves significantly by 2.8 points to 45.7\%. This demonstrates that enhancing deep semantic features while preserving shallow spatial details is critical. Notably, HFESNet also reduces GFLOPS by 7.7\% and parameters by 22.4\%.
Next, we progressively add the other modules. The introduction of ESOP yields a +0.5 point gain in AP$_{50}$, confirming its ability to effectively aggregate features for small objects. The GAPE module further boosts AP$_{50}$ by 1.1 points to 47.3\%, which we attribute to its geometrically-aware encoding that enhances discrimination of dense instances. Finally, the SQR training strategy adds another 0.7 points to AP$_{50}$ without any computational overhead by reinforcing supervision in later decoder stages.
The full HEDS-DETR model achieves an AP$_{50}$ of 48.0\% and an AP$_S$ of 20.9\%, marking a substantial improvement of +5.1 and +3.3 points over the baseline, respectively. Crucially, it maintains a real-time inference speed of 131.6 FPS, striking an excellent balance between accuracy and efficiency. These results systematically confirm that each component synergistically contributes to the model's superior performance in small object detection.

\subsection{Comparative Analysis of Different ESOP Schemes}

\begin{table*}
  \centering
  \caption{Comparison of Different ESOP Variants on VisDrone.}
  \label{tab:soep-ablation}
  % 将表格宽度缩放至 \textwidth，高度等比例缩放
  \scalebox{0.8}{
  \resizebox{\textwidth}{!}{%
    \begin{tabular}{@{}>{\RaggedRight\arraybackslash}p{2.5cm}
                     *{4}{S[table-format=2.1]}
                     S[table-format=2.1]
                     S[table-format=2.2]@{}}
      \toprule
      \textbf{Variant} & \multicolumn{4}{c}{\textbf{Average Precision}} & \textbf{GFLOPS} & \textbf{Params} \\
      \cmidrule(lr){2-5}
      & \multicolumn{1}{c}{AP$_{50}$ (\%)} 
      & \multicolumn{1}{c}{AP$_{50\text{-}90}$ (\%)} 
      & \multicolumn{1}{c}{AP$_{s}$ (\%)} 
      & \multicolumn{1}{c}{AP$_{m}$ (\%)} 
      & & {(M)} \\
      \midrule
      ESOP       & \textbf{44.6} & \textbf{26.7} & \textbf{18.3} & \textbf{37.0} & 65.2 & 20.50 \\
      ESOP-Conv   & 43.7 & 26.0 & 18.2 & 35.3 & 59.7 & 20.05 \\
      ESOP-NP & 43.1 & 25.5 & 18.1 & 34.7 & 62.4 & 20.28 \\
      RT-DETR-R18   & 42.9 & 25.6 & 17.6 & 35.4 & \textbf{57.0} & \textbf{19.88} \\
      \bottomrule
    \end{tabular}%
  }
  }
  \vspace{0.5em}
\end{table*} 

We conduct an ablation study to analyze the design choices within our ESOP module, with results presented in \autoref{tab:soep-ablation}. Our full ESOP architecture consistently outperforms the baseline and its variants, achieving an AP$_{50}$ of 44.6\%. This marks a 1.7-point improvement over the RT-DETR-R18 baseline, demonstrating a substantial accuracy gain for a modest increase in computational cost (+8.2 GFLOPS).
The efficacy of ESOP's core components is validated by analyzing its degraded variants. First, replacing the Space-to-depth convolution (SpdConv) with a standard strided convolution (ESOP-Conv) causes a significant 0.9-point drop in AP$_{50}$. This confirms our hypothesis that SpdConv is crucial for preserving the fine-grained spatial information necessary for small object detection during downsampling. Second, removing the COKBlock pre-fusion module (ESOP-NP) leads to an even larger performance degradation, with AP$_{50}$ dropping by 1.5 points. This underscores the COKBlock's vital role in effectively integrating cross-scale features that have significant semantic disparities. 
Collectively, these findings demonstrate that both the SpdConv-based feature preservation and the COKBlock-based fusion are indispensable to ESOP's effectiveness, justifying its design for superior detection performance at a marginal computational overhead.

\subsection{SQR Validation Experiments}

\begin{table*}
  \centering
  \caption{Comparison of Different SQR Variants VisDrone.}
  \label{tab:sqr-ablation}
  \resizebox{\textwidth}{!}{%
    \begin{tabular}{@{}>{\RaggedRight\arraybackslash}p{2.5cm}
                     *{4}{S[table-format=2.1]}
                     c
                     S[table-format=1.0]
                     S[table-format=2.1]@{}}
      \toprule
      \textbf{Model}
      & \multicolumn{4}{c}{\textbf{Average Precision}}
      & \textbf{Train Time}
      & \textbf{Loss Queries}
      & \textbf{Loss Queries} \\
      \cmidrule(lr){2-5}
      & \multicolumn{1}{c}{AP$_{50}$ (\%)}
      & \multicolumn{1}{c}{AP$_{50\text{-}90}$ (\%)}
      & \multicolumn{1}{c}{AP$_{s}$ (\%)}
      & \multicolumn{1}{c}{AP$_{m}$ (\%)}
      & 
      &  {(D$^2$)}
      &  {(D$^3$)}  \\
      \midrule
      SQR-Baseline       & 47.3 & 28.6 & 20.7 & 38.6 & 1.00× & 1 & 1 \\
      SQR I              & 47.3 & 28.6 & 20.6 & 39.1 & 1.06× & 1 & 2 \\
      SQR II             & \textbf{48.0} & \textbf{29.4} & \textbf{20.9} & \textbf{40.2} & 1.34× & 2 & 3 \\
      SQR III            & 47.4 & 28.6 & 20.6 & 39.2 & 1.40× & 2 & 4 \\
      \bottomrule
    \end{tabular}%
  }
\end{table*} 

We conducted experiments to validate the design of our Sequential Query Reinforcement (SQR) strategy. As presented in Table \ref{tab:sqr-ablation}, we compare three SQR variants against a strong baseline (HEDS-DETR without SQR). SQR II emerges as the optimal configuration, significantly boosting AP$_{50}$ by 0.7 points to 48.0\% and AP by 0.8 points to 29.4\%. This performance gain stems from its incremental supervision design, which effectively alleviates decoder cascading errors and stabilizes box optimization. While SQR II increases training time by 34\%, this is a justifiable trade-off for the substantial accuracy improvement at no extra inference cost. In contrast, SQR I yields negligible improvement, indicating that minimal supervision is insufficient. SQR III offers only marginal gains at a higher training cost (1.40x), suggesting that its query design introduces a disruptive learning gap.

\begin{table}[h]
    \centering
    \setlength{\tabcolsep}{2.5pt} % 减小列间距
    \begin{minipage}{0.50\textwidth}
        \resizebox{0.98\linewidth}{!}{%
            \begin{tabular}{l|c|c|c}
                \toprule
                \textbf{Model} & \textbf{TP Threshold} & \textbf{TP F Rate (\%)} & \textbf{FP E Rate (\%)} \\
                \midrule
                SQR-Baseline     & IoU \textgreater \ 0.25 & 31.91 & 72.40 \\
                SQR II           & IoU \textgreater \ 0.25 & \textbf{30.79} & \textbf{69.42} \\
                \midrule
                SQR-Baseline     & IoU \textgreater \ 0.50 & 32.26 & 72.38 \\
                SQR II           & IoU \textgreater \ 0.50 & \textbf{28.35} & \textbf{69.21} \\
                \midrule
                SQR-Baseline     & IoU \textgreater \ 0.75 & 27.57 & 72.32 \\
                SQR II           & IoU \textgreater \ 0.75 & \textbf{22.36} & \textbf{69.44} \\
                \bottomrule
            \end{tabular}%
        }
        \caption{Comparative Analysis of Localization Stability Under Varying IoU Thresholds.}
        \label{tab:sqr-tpfp-contrast}
    \end{minipage}
\end{table}

To further analyze localization stability, we evaluated the True Positive Fading Rate (TP F Rate) and False Positive Exacerbation Rate (FP E Rate) under varying IoU thresholds, as detailed in Table \ref{tab:sqr-tpfp-contrast}. SQR II consistently reduces both error rates compared to the baseline. Critically, the performance gap widens as the localization requirement becomes stricter. For instance, at a stringent IoU threshold of 0.75, SQR II slashes the TP F Rate by 5.21 points (from 27.57\% to 22.36\%) and the FP E Rate by 2.88 points. This trend strongly validates that our SQR strategy enhances bounding box precision by mitigating cascading errors, particularly for high-quality detections.

\subsection{Generalization Ability}

\begin{table}[h]
\centering
\setlength{\tabcolsep}{1.3pt} % 减小列间距
\caption{Model generalization proof experiment in CARPK dataset (\%).}
\label{tab:model_comparison_percent_booktabs}
\begin{tabular}{lcccc}
\toprule
\textbf{Model} & \textbf{AP(\%)} & \textbf{AP\textsubscript{50} (\%)} & \textbf{AP\textsubscript{s} (\%)} & \textbf{AP\textsubscript{m} (\%)} \\
\midrule

YOLOv8m  & 55.2 & 80.0 & 8.8  & 56.6 \\
YOLOv9m  & 54.0 & 79.9 & 9.5  & 55.4 \\
YOLOv10m & 53.8 & 78.9 & \textbf{10.0} & 55.3 \\
YOLOv11m & 54.8 & 80.9 & 9.2  & 56.3 \\
YOLOv12m & 54.8 & 80.0 & 9.9  & 56.3 \\
RT-DETR-R18      & 48.6 & 79.8 & 9.4  & 49.9 \\
DFINE    & 51.1 & 78.5 & 7.3  & 52.7 \\
DEIM     & 52.5 & 78.7 & 9.2  & 53.9 \\
HEDS-DETR (Ours)   & \textbf{56.4} & \textbf{81.9} & \textbf{10.0} & \textbf{58.0} \\
\bottomrule
\end{tabular}
\end{table}

To evaluate the generalization ability of our model, we conducted experiments on the CARPK dataset, a small-scale vehicle detection benchmark. This tests the model's adaptability to a different data domain, where Transformer-based architectures traditionally struggle due to their data-hungry nature compared to CNN-based models.
As shown in Table \ref{tab:model_comparison_percent_booktabs}, our HEDS-DETR defies this convention. It achieves state-of-the-art performance with 56.4\% AP and 81.9\% AP$_{50}$, surpassing not only other DETR-based methods (e.g., +5.3 AP over DFINE) but also the top-performing CNN-based model, YOLOv8m. Notably, HEDS-DETR also matches the best performance on small objects (AP$_s$) with a score of 10.0\%. These results demonstrate that our proposed components, particularly the feature-enhancing backbone, effectively improve the model's learning efficiency and grant it strong generalization capabilities, even on limited data.

\subsection{Visualization}
To provide an intuitive understanding of the localization improvements, we visualize the sampling behavior of the deformable cross-attention in the final decoder layer in Figure \ref{fig:deform-att}. These visualizations depict the sampling points, $p_m$, and their associated attention weights, $a_m$. Each sampling point is determined by adding a learned offset $\Delta p_m$ to a query's reference point $Ref$, such that $p_m = Ref + \Delta p_m$. The color intensity of each point corresponds to its attention weight.
A clear qualitative difference emerges when comparing the baseline (middle row) with our HEDS-DETR (bottom row). Our model’s sampling points are tightly concentrated within the object’s foreground area, and the associated attention weights are significantly stronger. This demonstrates that our proposed strategies, SQR and GAPE, effectively guide the model to focus its attention on salient object regions. This enhanced focus leads to sharper boundary prediction and ultimately validates the superior localization accuracy of our method.

\begin{figure}
\centering
\includegraphics[scale=0.29]{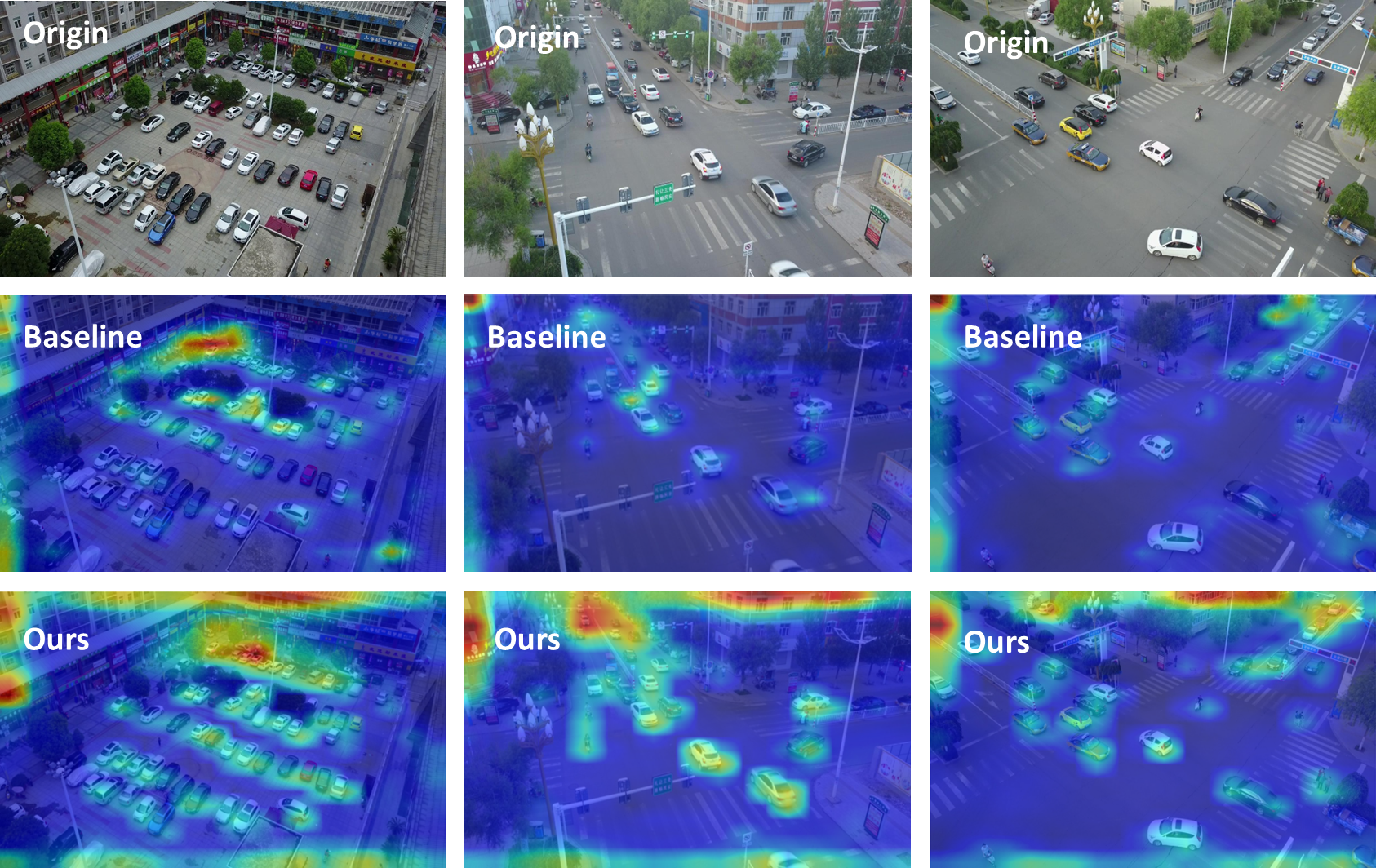}
\caption{Visualization of the attention maps in the AIFI \cite{zhao2024detrs} module.}\label{fig:att-map}
\end{figure}

To further investigate the feature enhancement capabilities of our HEGSNet backbone, we visualize the aggregated self-attention maps from the Intra-scale Feature Interaction (AIFI) module. The final attention map, $\mathbf{M}_{attn}$, is generated by averaging the raw self-attention weights across all attention heads and then across the query dimension to compute an importance score for each spatial token. This process is formally expressed as:

\begin{equation}
\mathbf{M}_{attn} = \text{Reshape}_{H,W}\left( \text{mean}_{j}\left( \frac{1}{N_h} \sum_{i=1}^{N_h} \mathbf{A}^{(i)}_{j,k} \right) \right)
\label{eq:attn_map}
\end{equation}

where $\mathbf{A}^{(i)}_{j,k}$ is the attention from query $j$ to key $k$ in head $i$. As depicted in Figure \ref{fig:att-map}, the resulting maps reveal a key qualitative difference. Our model, powered by HEGSNet, demonstrates a markedly superior ability to focus on dense clusters of small objects and their critical contextual cues compared to the baseline. This provides compelling visual evidence that our backbone architecture significantly enhances feature discriminability, a crucial factor for robust object localization and foreground-background separation in challenging scenes.

\section{Conclusion}\label{sec13}
This paper introduced HEDS-DETR, a new real-time object detector that effectively resolves the long-standing performance-efficiency trade-off in challenging aerial imagery. Our approach holistically enhances the standard DETR architecture by tackling key bottlenecks in both feature representation and localization. We designed a frequency-aware backbone (HFESNet) and a lightweight pyramid (ESOP) to preserve crucial small-object details, and refined the decoder's localization capability through a novel training strategy (SQR) and a geometry-aware encoding (GAPE).
Extensive experiments on the VisDrone dataset validate the superiority of our design. HEDS-DETR not only surpasses its RT-DETR baseline by a significant margin (+3.8\% AP and +5.1\% AP$_{50}$) but does so with approximately 17\% fewer parameters while maintaining real-time performance. This combination of high accuracy and efficiency confirms that our approach achieves a superior trade-off between performance and efficiency for object detection in aerial scenes, particularly excelling in dense and small-object scenarios.
Future work will focus on generalizing these architectural principles. We believe the concepts of frequency-aware feature extraction and geometry-informed decoding can be extended to advance other dense prediction tasks in aerial vision, such as instance segmentation and object tracking.

\section{Acknowledgments}
This research was supported by Natural Science Foundation of Guangdong Province (Grant No. 2025A1515011771) and Guangzhou Science and Technology Plan Project (Grant No. 2023B01J0046, 2024E04J1242).

\section{Data availability}
The datasets generated during or analysed during the current study are available from the corresponding author on reasonable request.

\bibliography{sn-bibliography.bib}

@article{wang2024spatial,
  title={A spatial arrangement preservation based stitching method via geographic coordinates of uav for farmland remote sensing image},
  author={Wang, Jiaxin and Du, Peng and Yang, Shuqin and Zhang, Zhitao and Ning, Jifeng},
  journal={IEEE Transactions on Geoscience and Remote Sensing},
  year={2024},
  publisher={IEEE}
}

@article{zheng2024efficient,
  title={An Efficient and Fast Image Mosaic Approach for Highway Panoramic UAV Images},
  author={Zheng, Haoxin and Chang, Zhanqiang and Li, Yakai and Zhu, Jie and Wang, Wei and Yang, Qing and Xie, Chou and Zhang, Jingfa and Liu, Jiaxi},
  journal={IEEE Journal of Selected Topics in Applied Earth Observations and Remote Sensing},
  year={2024},
  publisher={IEEE}
}

@article{henn2024surface,
  title={Surface heat monitoring with high-resolution UAV thermal imaging: Assessing accuracy and applications in urban environments},
  author={Henn, Katrina Ariel and Peduzzi, Alicia},
  journal={Remote Sensing},
  volume={16},
  number={5},
  pages={930},
  year={2024},
  publisher={MDPI}
}

@inproceedings{lin2014microsoft,
  title={Microsoft coco: Common objects in context},
  author={Lin, Tsung-Yi and Maire, Michael and Belongie, Serge and Hays, James and Perona, Pietro and Ramanan, Deva and Doll{\'a}r, Piotr and Zitnick, C Lawrence},
  booktitle={Computer vision--ECCV 2014: 13th European conference, zurich, Switzerland, September 6-12, 2014, proceedings, part v 13},
  pages={740--755},
  year={2014},
  organization={Springer}
}

@inproceedings{carion2020end,
  title={End-to-end object detection with transformers},
  author={Carion, Nicolas and Massa, Francisco and Synnaeve, Gabriel and Usunier, Nicolas and Kirillov, Alexander and Zagoruyko, Sergey},
  booktitle={European conference on computer vision},
  pages={213--229},
  year={2020},
  organization={Springer}
}

@article{tong2020recent,
  title={Recent advances in small object detection based on deep learning: A review},
  author={Tong, Kang and Wu, Yiquan and Zhou, Fei},
  journal={Image and Vision Computing},
  volume={97},
  pages={103910},
  year={2020},
  publisher={Elsevier}
}

@inproceedings{zhang2021vit,
  title={ViT-YOLO: Transformer-based YOLO for object detection},
  author={Zhang, Zixiao and Lu, Xiaoqiang and Cao, Guojin and Yang, Yuting and Jiao, Licheng and Liu, Fang},
  booktitle={Proceedings of the IEEE/CVF international conference on computer vision},
  pages={2799--2808},
  year={2021}
}

@inproceedings{zhu2021tph,
  title={TPH-YOLOv5: Improved YOLOv5 based on transformer prediction head for object detection on drone-captured scenarios},
  author={Zhu, Xingkui and Lyu, Shuchang and Wang, Xu and Zhao, Qi},
  booktitle={Proceedings of the IEEE/CVF international conference on computer vision},
  pages={2778--2788},
  year={2021}
}

@article{he2024key,
  title={Key technologies and applications of UAVs in underground space: a review},
  author={He, Bin and Ji, Xiangxin and Li, Gang and Cheng, Bin},
  journal={IEEE Transactions on Cognitive Communications and Networking},
  volume={10},
  number={3},
  pages={1026--1049},
  year={2024},
  publisher={IEEE}
}

@article{peng2024lgff,
  title={LGFF-YOLO: small object detection method of UAV images based on efficient local--global feature fusion},
  author={Peng, Hongxing and Xie, Haopei and Liu, Huanai and Guan, Xianlu},
  journal={Journal of Real-Time Image Processing},
  volume={21},
  number={5},
  pages={167},
  year={2024},
  publisher={Springer}
}

@inproceedings{zhao2024detrs,
  title={Detrs beat yolos on real-time object detection},
  author={Zhao, Yian and Lv, Wenyu and Xu, Shangliang and Wei, Jinman and Wang, Guanzhong and Dang, Qingqing and Liu, Yi and Chen, Jie},
  booktitle={Proceedings of the IEEE/CVF conference on computer vision and pattern recognition},
  pages={16965--16974},
  year={2024}
}

@inproceedings{lin2017feature,
  title={Feature pyramid networks for object detection},
  author={Lin, Tsung-Yi and Doll{\'a}r, Piotr and Girshick, Ross and He, Kaiming and Hariharan, Bharath and Belongie, Serge},
  booktitle={Proceedings of the IEEE conference on computer vision and pattern recognition},
  pages={2117--2125},
  year={2017}
}

@inproceedings{tan2020efficientdet,
  title={Efficientdet: Scalable and efficient object detection},
  author={Tan, Mingxing and Pang, Ruoming and Le, Quoc V},
  booktitle={Proceedings of the IEEE/CVF conference on computer vision and pattern recognition},
  pages={10781--10790},
  year={2020}
}

@inproceedings{liu2018path,
  title={Path aggregation network for instance segmentation},
  author={Liu, Shu and Qi, Lu and Qin, Haifang and Shi, Jianping and Jia, Jiaya},
  booktitle={Proceedings of the IEEE conference on computer vision and pattern recognition},
  pages={8759--8768},
  year={2018}
}

@inproceedings{sun2021sparse,
  title={Sparse r-cnn: End-to-end object detection with learnable proposals},
  author={Sun, Peize and Zhang, Rufeng and Jiang, Yi and Kong, Tao and Xu, Chenfeng and Zhan, Wei and Tomizuka, Masayoshi and Li, Lei and Yuan, Zehuan and Wang, Changhu and others},
  booktitle={Proceedings of the IEEE/CVF conference on computer vision and pattern recognition},
  pages={14454--14463},
  year={2021}
}

@inproceedings{chen2023enhanced,
  title={Enhanced training of query-based object detection via selective query recollection},
  author={Chen, Fangyi and Zhang, Han and Hu, Kai and Huang, Yu-Kai and Zhu, Chenchen and Savvides, Marios},
  booktitle={Proceedings of the IEEE/CVF conference on computer vision and pattern recognition},
  pages={23756--23765},
  year={2023}
}

@inproceedings{liu2022petr,
  title={Petr: Position embedding transformation for multi-view 3d object detection},
  author={Liu, Yingfei and Wang, Tiancai and Zhang, Xiangyu and Sun, Jian},
  booktitle={European conference on computer vision},
  pages={531--548},
  year={2022},
  organization={Springer}
}

@article{liu2022dab,
  title={Dab-detr: Dynamic anchor boxes are better queries for detr},
  author={Liu, Shilong and Li, Feng and Zhang, Hao and Yang, Xiao and Qi, Xianbiao and Su, Hang and Zhu, Jun and Zhang, Lei},
  journal={arXiv preprint arXiv:2201.12329},
  year={2022}
}

@article{huang2024discriminative,
  title={Discriminative features enhancement for low-altitude UAV object detection},
  author={Huang, Shuqin and Ren, Shasha and Wu, Wei and Liu, Qiong},
  journal={Pattern Recognition},
  volume={147},
  pages={110041},
  year={2024},
  publisher={Elsevier}
}

@article{gao2024oriented,
  title={An Oriented Ship Detection Method of Remote Sensing Image with Contextual Global Attention Mechanism and Lightweight Task-Specific Context Decoupling},
  author={Gao, Gui and Wang, Yajun and Chen, Yuhao and Yang, Gang and Yao, Libo and Zhang, Xi and Li, Hengchao and Li, Gaosheng},
  journal={IEEE Transactions on Geoscience and Remote Sensing},
  year={2024},
  publisher={IEEE}
}

@article{zhu2021detection,
  title={Detection and tracking meet drones challenge},
  author={Zhu, Pengfei and Wen, Longyin and Du, Dawei and Bian, Xiao and Fan, Heng and Hu, Qinghua and Ling, Haibin},
  journal={IEEE Transactions on Pattern Analysis and Machine Intelligence},
  volume={44},
  number={11},
  pages={7380--7399},
  year={2021},
  publisher={IEEE}
}

@article{ren15fasterrcnn,
    Author = {Shaoqing Ren and Kaiming He and Ross Girshick and Jian Sun},
    Title = {{Faster R-CNN}: Towards Real-Time Object Detection with Region Proposal Networks},
    Journal = {arXiv preprint arXiv:1506.01497},
    Year = {2015}
}

@inproceedings{cai2018cascade,
  title={Cascade r-cnn: Delving into high quality object detection},
  author={Cai, Zhaowei and Vasconcelos, Nuno},
  booktitle={Proceedings of the IEEE conference on computer vision and pattern recognition},
  pages={6154--6162},
  year={2018}
}

@inproceedings{zhang2020bridging,
  title     =  {Bridging the Gap Between Anchor-based and Anchor-free Detection via Adaptive Training Sample Selection},
  author    =  {Zhang, Shifeng and Chi, Cheng and Yao, Yongqiang and Lei, Zhen and Li, Stan Z.},
  booktitle =  {CVPR},
  year      =  {2020}
}

@inproceedings{feng2021tood,
    title={TOOD: Task-aligned One-stage Object Detection},
    author={Feng, Chengjian and Zhong, Yujie and Gao, Yu and Scott, Matthew R and Huang, Weilin},
    booktitle={ICCV},
    year={2021}
}

@misc{peng2024dfineredefineregressiontask,
      title={D-FINE: Redefine Regression Task in DETRs as Fine-grained Distribution Refinement}, 
      author={Yansong Peng and Hebei Li and Peixi Wu and Yueyi Zhang and Xiaoyan Sun and Feng Wu},
      year={2024},
      eprint={2410.13842},
      archivePrefix={arXiv},
      primaryClass={cs.CV},
      url={https://arxiv.org/abs/2410.13842}, 
}

@misc{huang2024deim,
      title={DEIM: DETR with Improved Matching for Fast Convergence},
      author={Shihua Huang and Zhichao Lu and Xiaodong Cun and Yongjun Yu and Xiao Zhou and Xi Shen},
      booktitle={Proceedings of the IEEE/CVF Conference on Computer Vision and Pattern Recognition},
      year={2025},
}

@misc{hsieh2017dronebasedobjectcountingspatially,
      title={Drone-based Object Counting by Spatially Regularized Regional Proposal Network}, 
      author={Meng-Ru Hsieh and Yen-Liang Lin and Winston H. Hsu},
      year={2017},
      eprint={1707.05972},
      archivePrefix={arXiv},
      primaryClass={cs.CV},
      url={https://arxiv.org/abs/1707.05972}, 
}

@article{zhu2020deformable,
  title={Deformable detr: Deformable transformers for end-to-end object detection},
  author={Zhu, Xizhou and Su, Weijie and Lu, Lewei and Li, Bin and Wang, Xiaogang and Dai, Jifeng},
  journal={arXiv preprint arXiv:2010.04159},
  year={2020}
}

@inproceedings{tang2024hic,
  title={HIC-YOLOv5: Improved YOLOv5 for small object detection},
  author={Tang, Shiyi and Zhang, Shu and Fang, Yini},
  booktitle={2024 IEEE international conference on robotics and automation (ICRA)},
  pages={6614--6619},
  year={2024},
  organization={IEEE}
}

@inproceedings{yang2022querydet,
  title={QueryDet: Cascaded sparse query for accelerating high-resolution small object detection},
  author={Yang, Chenhongyi and Huang, Zehao and Wang, Naiyan},
  booktitle={Proceedings of the IEEE/CVF Conference on computer vision and pattern recognition},
  pages={13668--13677},
  year={2022}
}

@inproceedings{xu2023dynamic,
  title={Dynamic coarse-to-fine learning for oriented tiny object detection},
  author={Xu, Chang and Ding, Jian and Wang, Jinwang and Yang, Wen and Yu, Huai and Yu, Lei and Xia, Gui-Song},
  booktitle={Proceedings of the IEEE/CVF Conference on Computer Vision and Pattern Recognition},
  pages={7318--7328},
  year={2023}
}

@inproceedings{yang2019clustered,
  title={Clustered object detection in aerial images},
  author={Yang, Fan and Fan, Heng and Chu, Peng and Blasch, Erik and Ling, Haibin},
  booktitle={Proceedings of the IEEE/CVF international conference on computer vision},
  pages={8311--8320},
  year={2019}
}

@misc{redmon2016lookonceunifiedrealtime,
      title={You Only Look Once: Unified, Real-Time Object Detection}, 
      author={Joseph Redmon and Santosh Divvala and Ross Girshick and Ali Farhadi},
      year={2016},
      eprint={1506.02640},
      archivePrefix={arXiv},
      primaryClass={cs.CV},
      url={https://arxiv.org/abs/1506.02640}, 
}

@inproceedings{wang2020cspnet,
  title={CSPNet: A new backbone that can enhance learning capability of CNN},
  author={Wang, Chien-Yao and Liao, Hong-Yuan Mark and Wu, Yueh-Hua and Chen, Ping-Yang and Hsieh, Jun-Wei and Yeh, I-Hau},
  booktitle={Proceedings of the IEEE/CVF conference on computer vision and pattern recognition workshops},
  pages={390--391},
  year={2020}
}

@inproceedings{freqformer2024,
    title={FreqFormer: Frequency-aware Transformer for Lightweight Image Super-resolution},
    author={Tao Dai and Jianping Wang and Hang Guo and Jinmin Li and Jinbao Wang and Zexuan Zhu},
    booktitle={IJCAI},
    year={2024}
}

@inproceedings{sunkara2022no,
  title={No more strided convolutions or pooling: A new CNN building block for low-resolution images and small objects},
  author={Sunkara, Raja and Luo, Tie},
  booktitle={Joint European conference on machine learning and knowledge discovery in databases},
  pages={443--459},
  year={2022},
  organization={Springer}
}

@inproceedings{cui2024omni,
  title={Omni-kernel network for image restoration},
  author={Cui, Yuning and Ren, Wenqi and Knoll, Alois},
  booktitle={Proceedings of the AAAI conference on artificial intelligence},
  volume={38},
  number={2},
  pages={1426--1434},
  year={2024}
}
\end{document}